\pdfoutput=1

\documentclass[11pt]{article}

\usepackage{acl}

\usepackage{times}
\usepackage{latexsym}
\usepackage{booktabs}
\usepackage{graphicx}
\usepackage[normalem]{ulem}
\useunder{\uline}{\ul}{}

\usepackage[T1]{fontenc}

\usepackage[utf8]{inputenc}

\usepackage{microtype}
\usepackage[most]{tcolorbox}
\usepackage{hyperref}
\usepackage{natbib}

\usepackage{inconsolata}
\usepackage{xcolor}

\usepackage{geometry} 
\usepackage{xcolor}
\usepackage{tcolorbox}
\usepackage{listings}
\usepackage{rotating}
\usepackage{placeins}

\title{Towards Interpretable Hate Speech Detection using Large Language Model-extracted Rationales}

\author{Ayushi Nirmal$^*$  ~~ Amrita Bhattacharjee$^*$ ~~ Paras Sheth ~~ Huan Liu  \\
        School of Computing and Augmented Intelligence \\ 
        Arizona State University\\ 
        \texttt{\{anirmal1, abhatt43, psheth5, huanliu\}@asu.edu}}

\begin{document}
\maketitle
\def\thefootnote{*}\footnotetext{These authors contributed equally to this work.}\def\thefootnote{\arabic{footnote}}
\begin{abstract}
Although social media platforms are a prominent arena for users to engage in interpersonal discussions and express opinions, the facade and anonymity offered by social media may allow users to spew hate speech and offensive content.
Given the massive scale of such platforms, there arises a need to automatically identify and flag instances of hate speech. 
Although several hate speech detection methods exist, most of these black-box methods are not interpretable or explainable by design. 
To address the lack of interpretability, in this paper, we propose to use state-of-the-art Large Language Models (LLMs) to extract features in the form of rationales from the input text, to train a base hate speech classifier, thereby enabling faithful interpretability by design.
Our framework effectively combines the textual understanding capabilities of LLMs and the discriminative power of state-of-the-art hate speech classifiers to make these classifiers faithfully interpretable.
Our comprehensive evaluation on a variety of English language social media hate speech datasets demonstrate: (1) the goodness of the LLM-extracted rationales, and (2) the surprising retention of detector performance even after training to ensure interpretability. 
All code and data will be made available at \url{https://github.com/AmritaBh/shield}.
\end{abstract}

\section{Introduction}


\begin{figure}[!ht]
    \centering
    \begin{tcolorbox}[colback=gray!10!white,colframe=black!50!white]
    \textbf{Content Warning:} This document contains content that some may find disturbing or offensive, including content that is discriminative, hateful, or violent in nature.
    \end{tcolorbox}
\end{figure}

Social media has become a platform of content sharing and discussions for a varied range of individuals with differing cultural and continental backgrounds. People use social media platforms to exchange information, and they frequently engage in dialectal conversations. These discussions are not always peaceful, they can degenerate into unpleasant altercations and bigoted arguments. Thus, social media platforms often become a host for hate speech. 
Hate speech is described as any deliberate and purposeful public communication meant to disparage a person or a group by expressing hatred, disdain, or contempt based on their social attributes (e.g., gender, race). In extreme cases, hate speech may often lead to real world harms such as hate crimes, for example the anti-Asian hate crimes during the COVID-19 pandemic~\cite{findling2022covid,han2023anti}. Therefore, it is essential to have automatic hate speech detection and moderation in place to maintain the integrity of social media platforms as well as to mitigate negative impacts in real-world scenarios such as increased violence towards minorities~\cite{laub2019hate}.

Given that the issue of hate speech on social media is a well-established problem, there have been several works to detect such online hate-speech ~\cite{schmidt2017survey,del2017hate}. While state of the art hate speech detection models have been able to achieve good performance on benchmark evaluation datasets, most of these models are built using transformer-based pre-trained language models or other deep neural network type models~\cite{sheth2023peace} that are not interpretable or explainable. However, the task of hate speech detection is a very sensitive task, and explainability of automated detectors is an essential and desirable feature. Model interpretability is essential not only for end-user understanding but also for understanding biased predictions, domain shifts, other errors in the prediction, etc.

While incorporating qualities of interpretability directly into deep neural network models such as pre-trained language model based detectors is challenging, one way to potentially perform this is by using an auxiliary model to provide explanations or rationales, that are subsequently used in training the detection model. This type of a method has been proposed and used in the FRESH framework ~\cite{jain2020learning}, where the authors use two disjoint networks, one for extracting the task-specific rationales, and then another that leverages those rationales to learn the classification task, thereby enabling faithful interpretability \textit{by construction}.

Inspired by this work, we propose a framework, where we use LLMs as the extractor model: we leverage the textual understanding and instruction-following capabilities of state-of-the-art LLMs to extract features from the input text, that is used to augment the training of a separate base hate speech detector, thereby facilitating faithful interpretability. Overall, our contributions in this paper are:

\begin{enumerate}
    \item We propose \textbf{SHIELD}, a framework that leverages LLM-extracted rationales to augment a base hate speech detection model to facilitate faithful interpretability.
    \item We evaluate the goodness of LLM-extracted features and rationales, and measure the alignment of such with human annotated rationales.
    \item Through comprehensive experiments on both implicit and explicit hate speech datasets, we show how \textbf{SHIELD} retains detection performance even after training with rationales for increased interpretability, despite the expected interpretability-accuracy trade-off.
\end{enumerate}



\section{Our SHIELD Framework}

\begin{figure}[!ht]
    \centering
    \includegraphics[width=\columnwidth]{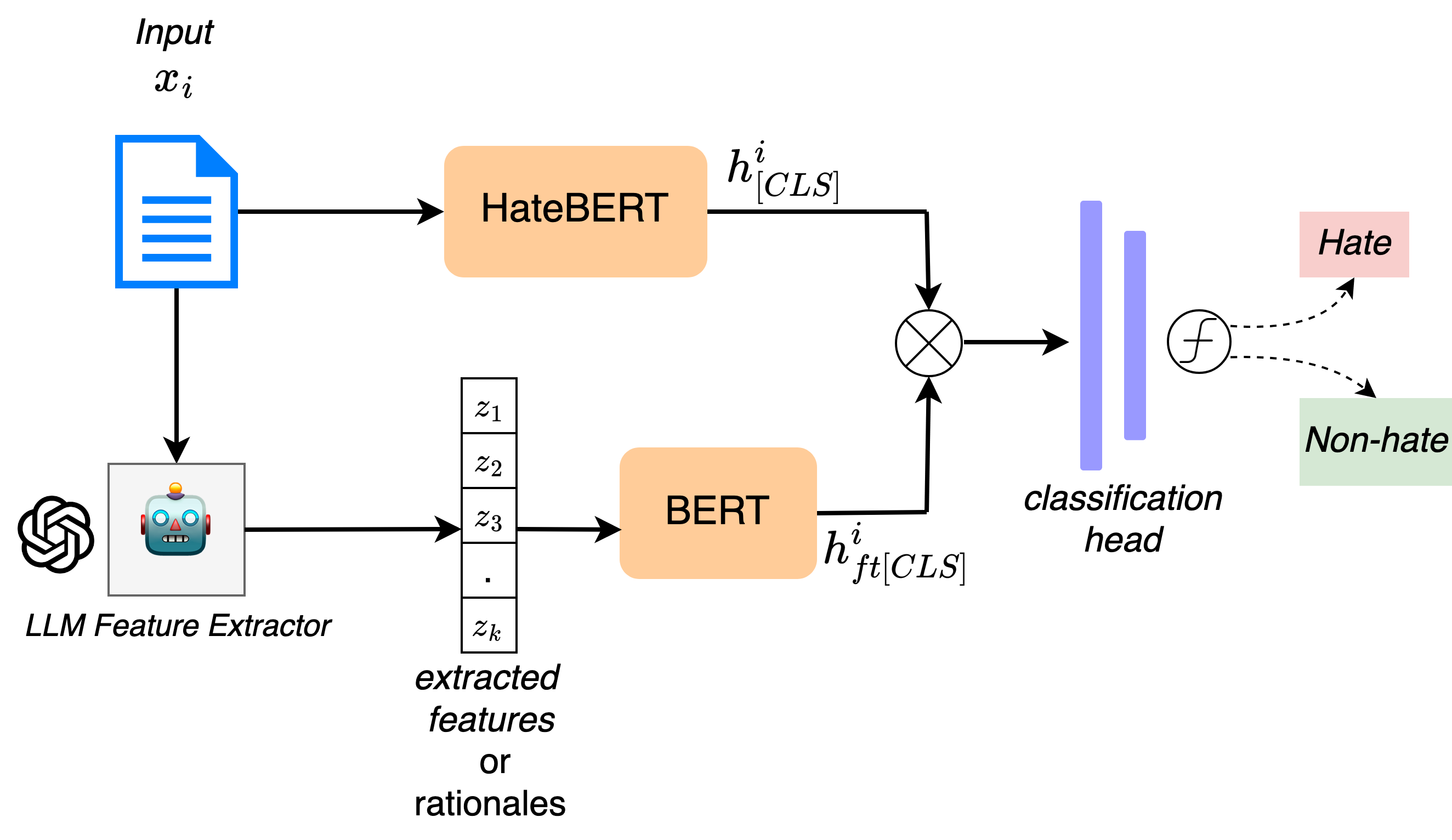}
    \caption{Our proposed \textbf{SHIELD} framework.}
    \label{fig:framework}
\end{figure}

We show our proposed \textbf{SHIELD} framework in Figure \ref{fig:framework}. In this section, we describe our framework in detail, elaborating on each of the components. 

\paragraph{LLM Feature Extractor} Our framework uses the state-of-the-art instruction-tuned large language models (LLMs) in an off-the-shelf manner as textual feature extractors. Although recent work has shown that LLMs struggle to perform the hate speech detection task ~\cite{li2023hot,zhu2023can} when used without any additional model or fine-tuning, we hypothesize that we can leverage the textual understanding capabilities of these LLMs to simply extract textual features in the form of rationales. Restricting the use of the LLM to a simple text-level task would ensure that such models are not directly being used for sensitive application tasks such as hate speech detection ~\cite{harrer2023attention}. 
For a given input text $x_i \in X$, we use our carefully designed task prompt to prompt the LLM to extract features from the text that promotes a hateful sentiment. In the context of explicit hate speech detection, such features could include categories such as derogatory words, cuss words, etc. Following similar work in ~\cite{bhattacharjee2023llms}, we also ask the LLM for rationales as to why the label is hateful or non-hateful. To perform this feature extraction, for each input text we prompt the LLM using the following prompt:

\vspace{1.5mm}
\noindent\fbox{%
    \parbox{0.95\columnwidth}{%
``You are a content moderation bot. Identify the list of rationales, list of derogatory language, list of cuss words that promote a hateful sentiment and respond with non-hateful if there are none. Note: The output should be in a json format.''
\\
Text: \textcolor{blue}{[input\_text]} \\

    }%
}
\vspace{1.5mm}

After post-processing the outputs, we have a list of $k$ textual features $\{z_j\}_{j=1}^{k}$ for the given input text $x_i$. 

\paragraph{Hate Speech Detector as Embedding Module} The next component in our framework is the base hate speech detector which we are trying to augment, such as HateBERT~\cite{caselli2020hatebert}. HateBERT is a BERT~\cite{devlin2018bert} model that is specifically fine-tuned on hate speech data. For each input text $x_i \in X$, instead of obtaining the labels or class probabilities, we take the last layer embedding of the \texttt{[CLS]} token, $h^{i}_{[CLS]}$, essentially containing all the information of the input text, that is relevant for the hate-speech detection task. 

\paragraph{Feature Embedding Model} For the textual features and rationales, $\{z_j\}_{j=1}^{k}$, we extracted via the LLM, we use a pre-trained transformer-based language model (PLM), such as BERT to embed these features. PLMs, even without any task-specific fine-tuning, provide rich, expressive latent representations for text. Therefore, we feed in the LLM-extracted textual features into a BERT (specifically, bert-base-uncased\footnote{https://huggingface.co/google-bert/bert-base-uncased}) model and obtain the last hidden layer embedding of the \texttt{[CLS]} token, and we denote this as $h^{i}_{ft[CLS]}$.

\paragraph{Embedding Fusion \& Classification} From the previous two components, for each input text $x_i$, we have two embeddings: text embedding $h^i_{[CLS]}$ from the base hate speech detector, and feature embedding $h^{i}_{ft[CLS]}$ from the feature embedding BERT model. To combine these two, we simply concatenate these embeddings: 

\begin{equation}
    h^{i}_{combined} = h^i_{[CLS]} \oplus h^{i}_{ft[CLS]}
\end{equation}

Note that while authors in ~\cite{jain2020learning} only use the extracted rationales in the subsequent detector model, we use a concatenated view in order to incorporate additional contextual features that may be very relevant to determining the hate or non-hate label~\cite{ocampo2023depth}. We then feed this combined embedding $h^{i}_{combined}$ into a feed-forward multi-layer perceptron with two fully connected layers and a ReLU activation~\cite{agarap2018deep} in between, to project it onto a smaller dimension space. Following previous work~\cite{pan2022improved,bhattacharjee2023conda}, we do this in order to retain important features and avoid overfitting of the model during training. We denote this MLP as $f(\cdot)$. Finally we compute the batch-wise binary cross entropy loss using the ground truth label $y_i$ for each input text $x_i$:


\begin{equation}
\label{eq1}
\begin{split}
    loss_{CE} & = -\frac{1}{n} \sum_i^n [\log p(y_i|f(h^{i}_{combined})) + \\
    & (1-y_i) \log (1-p(y_i|f(h^{i}_{combined}))]
\end{split}
\end{equation}

where $n$ is the batch size. Since we are using the BERT feature embedding model just to encode the textual features $z$, we keep this model frozen and train the remainder of the framework with this simple loss. 



\section{Methodology and Experimental Settings}


In this section, we discuss our methodology in detail including the datasets we included, the baseline models for hate speech detection along with the experimental settings. 

\subsection{Datasets}

In order to evaluate \textbf{SHIELD}, we use both explicit and implicit hate speech datasets. For explicit hate, we include publicly available benchmark datasets from the following social media platforms: \{GAB, Twitter, YouTube, and Reddit\}. All these datasets are in the English language. \textbf{GAB} ~\cite{mathew2021hatexplain} is a collection of annotated posts from the GAB website. It consists of binary labels indicating whether a post is hateful or not. \textbf{Reddit} ~\cite{kennedy2020constructing} is a collection of posts indicating whether it is hateful or not. \textbf{Twitter} ~\cite{mathew2021hatexplain} contains instances of hate speech gathered from tweets on the Twitter platform. Finally, \textbf{YouTube} ~\cite{salminen2018anatomy} is a collection of hateful expressions and comments posted on the YouTube platform. 
We further pre-process these according to the method followed in ~\cite{sheth2023causality}, in order to get cleaned binary labels.  A summary of the datasets and the distribution of hateful posts and non-hateful posts can be found in Table \ref{tab:data-stats}. 

We also include implicit hate speech in our evaluation: while subtle forms of abuse may not be perceived as overtly harmful initially, they nonetheless perpetuate similar degrees of damage over time owing to their covert nature. Therefore, the detection of implicit hate speech becomes even more important. For this reason, we evaluate our proposed model on the \textbf{Implicit Hate Speech Corpus} ~\cite{elsherief2018hate}. This dataset encompasses posts compiled from Twitter, annotated as either explicit hate, implicit hate, or non-hate speech. We exclusively utilize implicit hate and non-hate for our binary classification task.


\begin{table}[]
\centering
\resizebox{\columnwidth}{!}{%
\begin{tabular}{@{}cccc@{}}
\toprule
\textbf{Dataset} & \textbf{\# of Posts} & \begin{tabular}[c]{@{}c@{}}\textbf{\# of Hateful} \\ \textbf{Posts}\end{tabular} & \textbf{Hate \%} \\ \midrule
GAB     & 14,240 & 11,920 & 83.7 \\
Reddit  & 37,164 & 10,562 & 28.4 \\
Twitter & 10,457 & 3,933  & 37.6 \\
YouTube & 5,052  & 1,699  & 33.6 \\ 
Implicit HS & 20,391 & 7,100  & 34.8 \\\bottomrule
\end{tabular}%
}
\caption{Dataset statistics for explicit and implicit hate speech datasets comprising data from different social media platforms. }
\label{tab:data-stats}
\end{table}

\subsection{Baselines}
\label{sec:baselines}

We compare our proposed \textbf{SHIELD} framework to a variety of different baselines in order to understand the impact of the augmentation with rationales. We use the following well-known baseline hate speech detection models:

\textbf{HateBERT}: This is also the base model used in our framework. HateBERT ~\cite{caselli2020hatebert} uses over 1.5 million Reddit messages from suspended communities known for encouraging hate speech to fine-tune the BERT-base model. We further fine-tune HateBERT on each dataset and report the performance. 

\textbf{HateXplain}:  Similarly, we fine-tune the HateXplain~\cite{mathew2021hatexplain} model on each of our datasets and report the performance. HateXplain model is trained on hateful posts along with the target community, the rationales, and the portion of the post on which human annotators' labelling decision is based. 

\textbf{PEACE}: We further extend our comparison on PEACE~\cite{sheth2023peace} framework which uses Sentiment and Aggression Cues to detect the overall sentiment of the text. 

\textbf{CATCH}: Furthermore, we compare our model with CATCH ~\cite{sheth2023causality} framework which disentangles the input representations into invariant and platform-dependent features.

\textbf{ChatGPT-1shot}: Apart from these hate speech specific detection models, we also compare our framework with an off-the-shelf \textbf{GPT-3.5} model, to understand how well the LLM performs on the same datasets. We do this in a one-shot manner, i.e., by proving the task instruction along with an example input and ground truth label.

\subsection{Experimental Settings}

To implement our proposed \textbf{SHIELD} framework, we use PyTorch and the Huggingface Transformers library. As shown in Figure \ref{fig:framework}, our first component uses an off-the-shelf LLM to extract the features and rationales. Here, we use OpenAI's GPT-3.5 (specifically, GPT-3.5-turbo-0613)\footnote{or otherwise commonly referred to as `ChatGPT'}, since it has been experimented on a variety of NLP tasks with huge success~\cite{guo2024investigation}. We access this model via the OpenAI API. For feature/rationale extraction and generation, we set the temperature to 0.1 and top\_p to 1. For the Feature Embedding Model we use a pre-trained, frozen BERT (bert-base-uncased) and for the Hate Speech Detector we use a pre-trained HateBERT\footnote{https://huggingface.co/GroNLP/hateBERT} model. We use AdamW optimizer~\cite{kingma2014adam} with a learning rate of $2\times10^{-5}$. Model training was performed on two machines: one with an NVIDIA GP102 [TITAN Xp] GPU with 12 GB VRAM, and another with an NVIDIA A100 GPU with 40GB RAM. 
For all detection experiments, we use accuracy as the evaluation metric.

    

\section{Results and Discussion}

In this section we describe our experiments and elaborate on the experimental results. To explore the feasibility and effectiveness of our proposed \textbf{SHIELD} framework, we aim to answer the following research questions:

\begin{itemize}
    \item \textbf{RQ1:} How well does ChatGPT perform on our set of hate speech detection datasets?
    \item \textbf{RQ2:} Can we leverage recent state-of-the-art LLMs to extract features in the form of rationales, and do these rationales align with human judgement?
    \item \textbf{RQ3:} Can \textbf{SHIELD} effectively retain/improve performance of the hate speech detector while facilitating faithful interpretability?
\end{itemize}

\subsection{Performance of ChatGPT on the hate speech detection task}

Several recent works test whether Large Language Models have the potential to reproduce human annotated ground truth labels in social computing tasks \cite{zhu2023can}. However, even after extensive pre-training on a large corpus of datasets, where LLMs are expected to perform well in this task, this is not the case. To further evaluate this beyond what other recent works have shown, we carefully craft a one-shot prompt and prompt ChatGPT to classify the input text, given a labeled example in the prompt. The outcome of this prompt is a single label representing hateful text as label ``1" and non-hateful text as label ``0" as shown in Table \ref{tab:gpt-classification}. 
\\

\begin{table}[!ht]
\centering
\begin{tabular}{@{}p{0.3\linewidth}p{0.6\linewidth}@{}}
\toprule
\textbf{Text} & "always thought it astounding no policing agency dares do this to the armed muslim compounds all over the country" \\
\midrule
\textbf{Prompt} & \textit{You are a hate speech detection bot. Given a text, respond with the classification label towards given text with either hateful labeled as 1 or non-hateful labeled as 0. Note: It is essential to give classification for all the texts.} \\
& \textit{Text:} "always thought it astounding no policing agency dares do this to the armed muslim compounds all over the country" \\
\midrule
\textbf{ChatGPT} & \\
\textbf{response} & 1 \\
\bottomrule
\end{tabular}
\caption{Examples of input text, prompt and ChatGPT's response for a data sample from the Twitter dataset.}
\label{tab:gpt-classification}
\end{table}

\begin{table*}
\centering
\resizebox{\textwidth}{!}{%
\begin{tabular}{@{}lc|cccc|c@{}}
\toprule
& \textbf{HateBERT} & \textbf{HateXplain} & \textbf{PEACE} & \textbf{CATCH} & \textbf{ChatGPT-1shot} & \textbf{SHIELD (ours)} \\
\midrule
\textbf{GAB}                     & 96 & \textbf{97} & 91 & 82 & 85.39 & \uline{96.3} \\ 
\textbf{YouTube}                 & 71 & \uline{72} & \uline{72} & \textbf{79} & 58.34 & 70 \\
\textbf{REDDIT}                  & \uline{94} & 93 & 93 & 86 & 65.05 & \textbf{94.5} \\
\textbf{Twitter}                 & 56 & 60  & 31 & \textbf{78} & 60.09 & \uline{64} \\
\textbf{Implicit HS}    & \textbf{78} & \uline{76} & 64 & -- & 65.68 & \textbf{78} \\
\bottomrule
\end{tabular}%
}
\caption{Evaluation results (test set accuracy) for our \textbf{SHIELD} framework vs. the baseline models. Implicit HS refers to the Implicit Hate Speech Corpus. Values in \textbf{bold} denote the best performance, and \underline{underlined} values denotes the second-best performance.}
\label{tab:results-table}
\end{table*}

We perform this classification using ChatGPT for all 5 datasets and compute the accuracy. We compare the results of this one-shot classification task with the baseline models (as described in Section \ref{sec:baselines}) and show the results in Table \ref{tab:results-table}. We see a stark difference in the performance of the baseline models vs. ChatGPT-1shot classification accuracies. While performance on the GAB dataset is satisfactory, ChatGPT struggles with the other 4 datasets with \textasciitilde 58-65\% accuracy. Similar observations have been reported in other recent work that have investigated the off-the-shelf performance of LLMs in hate speech detection~\cite{li2023hot,zhu2023can}. 

While this shows ChatGPT and possibly other LLMs struggle at hate speech detection when used as a detector directly, these models have also been shown to have impressive textual understanding capabilities. Perhaps, simply using these models to extract features or rationales, instead of performing the entire detection task, might be beneficial. We evaluate this in the following subsection. 


\subsection{Goodness of ChatGPT extracted features or rationales}




We are interested to evaluate the textual and contextual understanding capabilities of ChatGPT in order to extract features in the form of rationales from the input text that are meaningful to the task of hate speech detection. Following a similar construction as in ~\cite{jain2020learning}, we use the LLM (i.e., GPT-3.5) as the \textit{extractor} model, which unlike the extractor model in ~\cite{jain2020learning}, does not require any additional task-specific fine-tuning. This is possible due to the instruction-following capabilities of recent LLMs. We carefully craft a prompt (as shown in Table \ref{tab:examples}) to extract \textit{cuss words}, \textit{derogatory language} and \textit{rationales} from the input text that serve as interpretable features that can be used in the subsequent \textit{predictor} model (HateBERT) in order to have a faithfully interpretable hate speech detector. In order to evaluate the goodness of the extracted features or rationales, we compare ChatGPT-extracted rationales with human-annotated ground truth rationales. We use the annotated rationale spans in the HateXplain~\cite{mathew2021hatexplain} dataset. After some standard pre-processing such as removing stop words, we compute the similarity between the ChatGPT extracted rationales for the input text from HateXplain dataset and the human-annotated rationales and report these scores in Table \ref{tab:sim-data}. We compute similarity metrics in both the token space (Jaccard and Overlap similarity) and in the latent space (Cosine and Semantic similarity with Universal Sentence Encoder embeddings~\cite{cer2018universal}) We see significant overlap and a high semantic similarity between the LLM and human rationales.  

\begin{figure*}[]
    \centering
    \includegraphics[width=\textwidth]{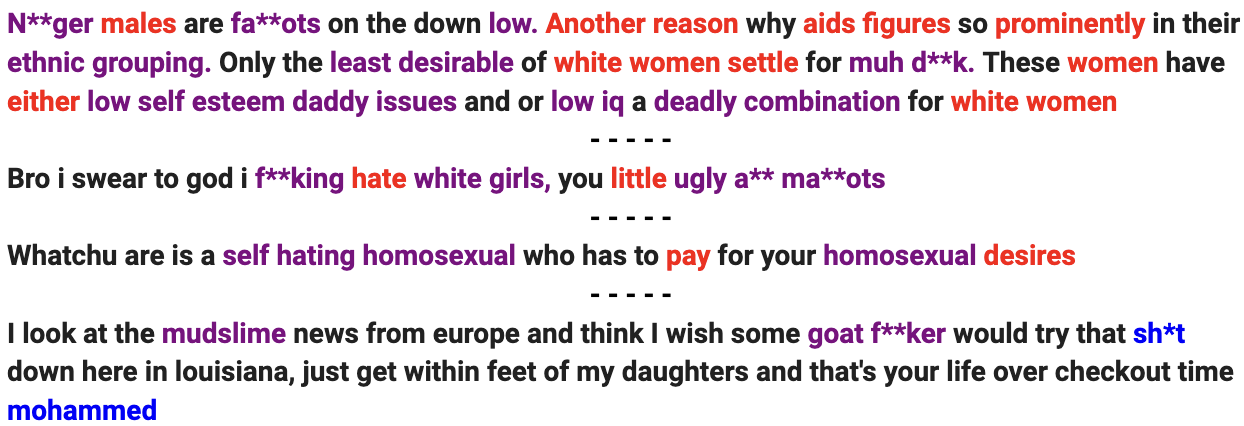}
    \caption{Examples with both \textcolor{blue}{LLM-annotated} and \textcolor{red}{human-annotated} rationales. Overlap is in \textcolor{purple}{purple}.}
    \label{fig:interpretability}
\end{figure*}

\begin{table*}[]
\centering
\resizebox{\textwidth}{!}{%
\begin{tabular}{@{}ccc@{}}
\toprule
\multicolumn{3}{c}{\begin{tabular}[c]{@{}c@{}}\textbf{Prompt:} \textit{You are a content moderation bot. Identify the list of rationales, list of derogatory language, list of cuss words that promote a} \\ \textit{hateful sentiment and respond with non-hateful if there are none. Note: The output should be in a json format. "Text"}\end{tabular}} \\ \midrule
\textbf{Dataset} &
  \textbf{Text} &
  \textbf{ChatGPT-extracted features and rationales} \\ \midrule
GAB &
  \begin{tabular}[c]{@{}c@{}}does this mean they arrested the mayor of london\\  have cordoned off every muzrat ghetto in the uk \\ every mosque closed down if not its business as usual\end{tabular} &
  \begin{tabular}[c]{@{}c@{}}\textcolor{blue}{arrested the mayor of london, cordoned off every muzrat} \\ \textcolor{blue}{ghetto in the uk, every mosque closed down}; \textcolor{red}{muzrat, ghetto} \end{tabular} \\ \midrule
REDDIT &
  \begin{tabular}[c]{@{}c@{}}"Shut up! 50 years ago we'd have you upside down with a \\ f**king fork up your a**! You can talk, you can talk, you can talk! \\ You're brave now motherf**ker! Throw his a** out, he's a n**ger! \\ HE'S A N**GER! HE'S A N**GER! A N**GER! \\ LOOK, THERE'S A N**GER!"\end{tabular} &
  \textcolor{red}{n**ger}; \textcolor{teal}{f**king, motherf**ker}\\ \midrule
Twitter &
  "show me the black muslims bc im calling bs on this and you should too" &
    \textcolor{blue}{black muslims}; \textcolor{red}{black, muslims} \\ \midrule
YouTube &
  "Dirty f**king israeli zionist jews" &
   \textcolor{red}{israeli, zionist, jews}; \textcolor{teal}{f**king} \\ \midrule
Implicit HS &
  \begin{tabular}[c]{@{}c@{}}"i'm so glad \#ferguson happened . you woke the world up \\ to how much black people hate white people ."\end{tabular} &
  \begin{tabular}[c]{@{}c@{}}\textcolor{blue}{ferguson, black people hate white people}; \\  \textcolor{red}{black people, white people}\end{tabular} \\ \bottomrule
\end{tabular}%
}
\caption{Examples from different datasets along with the LLM-extracted features and rationales. Rationales are in \textcolor{blue}{blue}, derogatory language is in \textcolor{red}{red}, cuss words are in \textcolor{teal}{teal}.}
\label{tab:examples}
\end{table*}

We present some examples from all 5 datasets in Table \ref{tab:examples}: the input text with a `hateful' label and the ChatGPT-extracted features. The three category of features are shown in different colors: \textcolor{blue}{rationales}, \textcolor{red}{derogatory language} and \textcolor{teal}{cuss words}. We see that the LLM is successfully able to identify the words and spans quite well. 

We also present some examples in Figure \ref{fig:interpretability} to qualitatively discern the overlap between the human-annotated rationales and the LLM-extracted ones. Text in \textcolor{red}{red} are rationales annotated by human annotators, text in \textcolor{blue}{blue} are rationales or words identified by the LLM and text in \textcolor{purple}{purple} are the spans where both the LLM and human annotations overlap. From these examples, we see that there is overall a high degree of overlap, and the LLM is able to capture semantically relevant portions of the text. Interestingly, we also see that while human annotators often annotate words or spans with lesser relevance to the task, the LLM extracted rationales do not contain these spans (such as `\textcolor{red}{aids figures}' and `\textcolor{red}{prominently}' in the first example in Figure \ref{fig:interpretability}). Using LLM-extracted rationales for training might be even more useful in such cases since some of the noisy signals in the data can be avoided.

\subsection{Hate speech detector performance after training with extracted rationales}

In this experiment, we try to train a hate speech detector with the extracted rationales additionally incorporated into the input text, to facilitate faithfully interpretable classifications. For this we use a HateBERT model as the base hate speech detector model and report results in Table \ref{tab:results-table}, along with results from other baselines. We see that our \textbf{SHIELD} framework performs at par with a simple HateBERT fine-tuned on the same dataset, i.e., at par with the base model. This performance retention is encouraging, since models are otherwise known to trade-off accuracy for interpretability ~\cite{dziugaite2020enforcing,bertsimas2019price}. Interestingly, in the Twitter dataset, we also see a significant $12.5\%$ performance jump by our \textbf{SHIELD} model as compared to the fine-tuned HateBERT model. This potentially might be due to noise in the Twitter dataset: the extracted rationales may provide more discriminative training signals thus allowing the detector to train on robust features instead of noisy ones, although more analysis is required to verify this claim.

\begin{table}[]
\centering
\resizebox{\columnwidth}{!}{%
\begin{tabular}{@{}cc@{}}
\toprule
\textbf{Similarity Metric} & \textbf{Similarity Coefficients (\%)} 
\\ \midrule
Jaccard Similarity  & 60.39 \\
Overlap Similarity  & 99.17 \\
Cosine Similarity  &  74.51\\ 
Semantic Similarity (via USE) & 56.09 \\
\bottomrule
\end{tabular}%
}
\caption{Similarity between HateXplain human explanations and LLM-extracted features/rationales.}
\label{tab:sim-data}
\end{table}


For some additional analysis on the effect of the framework components, we modify the choice of the base pre-trained language models in the two model components: the hate speech detector, and  the feature extractor. The specific variations we experiment with are: (1) the original \textbf{SHIELD} framework which has HateBERT as the hate speech detector (HSD) and bert-base-uncased as the feature embedding model (FE), (2) \textbf{SHIELD} with a pre-trained roberta-base as the HSD instead of HateBERT and (3) \textbf{SHIELD} with a pre-trained roberta-base as the FE instead of bert-base-uncased. We choose to perform this analysis with roberta instead of the two bert based models since RoBERTa~\cite{liu2019roberta} has been shown to sometimes have better performance than BERT~\cite{devlin2018bert} on a variety of natural language understanding tasks~\cite{tarunesh2021trusting}. We report the results of this analysis in Table \ref{tab:ablation}. Overall, we see some variation in performance on the model choice for the HSD and FE components. While roberta-base as the FE component marginally helps to improve performance for only one dataset, i.e., GAB, roberta-base as the HSD instead of HateBERT achieves higher performance for three datasets. This is particularly interesting since, unlike HateBERT, the pre-trained roberta-base is not specifically trained on the hate speech task. 

Overall, \textbf{SHIELD} shows promising results in leveraging LLM-extracted rationales into augmenting a base hate speech detector, to facilitate faithful interpretability, while maintaining detection performance.

\begin{table*}
\centering
\resizebox{\textwidth}{!}{%
\begin{tabular}{@{}lcccccc@{}}
\toprule
& \textbf{GAB} & \textbf{YouTube} & \textbf{REDDIT} & \textbf{Twitter} & \textbf{Implicit HS} \\
\midrule
\textbf{SHIELD (roberta-base HSD)}                 & 87.53 & \textbf{72.2} & 84.8 & \textbf{67.03} & \textbf{78.36} \\
\textbf{SHIELD (roberta-base FE)}   & \textbf{96.42} & 69.27 & 94.21 & 56.22 & 77.52 \\
\textbf{SHIELD} & 96.3 & 70 & \textbf{94.5} & 64 & 78 \\
\bottomrule
\end{tabular}%
}
\caption{Analysis of HSD and FE model choices in the \textbf{SHIELD} framework. HSD: hate speech detector, FE: feature embeddeing model. The original \textbf{SHIELD} framework has HateBERT as the hate speech detector and bert-base-uncased as the feature embedding model. Numbers in \textbf{bold} denote best performaning model variant for each dataset.}
\label{tab:ablation}
\end{table*}

\section{Related Work}


\subsection{Hate Speech Detection}
There are two primary methods for approaching the detection of hate speech. Leveraging new or supplementary data is the first strategy. This involves making advantage of user attributes~\cite{del2023socialhaterbert}, dataset annotator features~\cite{yin2022annobert}, or comprehending the ramifications of hateful posts~\cite{kim2022generalizable}. One study, for instance, used the consequences of hateful posts to train a model on contrastive pairs that represent hate content in order to detect implicit hate speech~\cite{kim2022generalizable}. An additional study~\cite{yin2022annobert} brought to light the challenge of reaching agreement among annotators on subjective issues such as recognizing hate speech, and it recommended that definitive labels and annotator traits be included in training to improve the efficacy of detection. In a different study~\cite{del2023socialhaterbert}, data from users' social situations and characteristics were analyzed to predict user satisfaction. But the problem with these strategies is that they could be challenging as access to auxiliary information across different platforms is seldom available.

The second tactic makes use of language models like BERT, which have been trained on large text datasets and are renowned for their capacity for generalization. The efficacy of these algorithms can be increased by fine-tuning them using particular hate speech datasets~\cite{caselli2020hatebert,mathew2021hatexplain}. One such example is HateBERT~\cite{caselli2020hatebert}, a model that was refined using over 1.6 million hostile remarks from Reddit and based on a BERT model. In a similar vein, HateXplain~\cite{mathew2021hatexplain} is another model created to recognize and interpret hate speech. Other strategies include concentrating on lexical indications~\cite{schmidt2017survey} such as POS tags used~\cite{markov2021exploring}, facial expressions, content-related portions of speech, or important phrases that communicate hate~\cite{elsherief2018hate}. In order to improve language model representations, one study manually determined that sentiment and hostility are causal cues~\cite{sheth2023peace}. Another study leveraged a causal graph to disentangle the input representations into platform specific (hate-target related features) and platform invariant features to enhance generalization capabilities for hate speech detection~\cite{sheth2023causality}. Although effective, this method also requires auxiliary data (such as hate target labels) which are seldom available across various platforms. 

\subsection{LLMs as Experts or Feature Extractors}

Recent advancements in LLM research have demonstrated improved performance across not only many natural language tasks~\cite{min2023recent}, but also more challenging domains such as writing and debugging code, performing mathematical reasoning~\cite{bubeck2023sparks}, etc. This has motivated a line of research where the community has been trying to evaluate how well these LLMs can perform different tasks. LLMs have shown promise in the task of data annotation~\cite{he2023annollm,bansal2023large}, information extraction ~\cite{dunn2022structured}, text classification~\cite{kocon2023chatgpt,bhattacharjee2024fighting}, and even reasoning ~\cite{ho2022large}. Given the ease with which these LLMs can be queried, these models often serve as faulty experts or pseudo oracles in many tasks. Past exploration has investigated whether language models can be used as factual knowledge bases~\cite{petroni2019language}. A recent work has explored the possibility of using LLMs in the hate speech detection task~\cite{kumarage2024harnessing}. Similar to our approach, authors in ~\cite{hasanain2023large} have tried to perform propaganda span annotation using language models. However, our approach focuses on leveraging the extracted spans, words and rationales to augment a detector model to enable interpretability in an otherwise black-box model.

\section{Conclusion and Future Work}

In this work, we explore the problem of hate speech detection on social media and propose a method to train interpretable classifiers using rationales extracted by large language models. Given the unsatisfactory performance of LLMs as off-the-shelf detectors for hate speech, we instead intend to leverage the textual understanding and instruction-following capabilities of LLMs such as ChatGPT to extract words and rationales from the text that are associated with the hate speech label. We propose a framework \textbf{SHIELD}, that uses these LLM-extracted rationales to augment the training of a base hate speech detector to facilitate it to be faithfully interpretable. We verify that the LLM-extracted rationales align with human judgement. We train and evaluate our framework on multiple benchmark datasets comprising both implicit and explicit hate speech from a variety of online social media platforms, and demonstrate how our \textbf{SHIELD} framework is able to maintain performance similar to the base model in spite of an expected accuracy-interpretability trade-off. Therefore, we have a faithfully interpretable hate speech detector that simply relies on LLM-extracted rationales instead of human-annotated.

While our work follows that of ~\cite{jain2020learning} and we establish faithfulness by construction, future work could explore better ways to evaluate the faithfulness of the resulting detector. In this work, we verified the goodness of the extracted rationales by comparing it with the ground truth for one dataset. Future work can investigate better automated ways to evaluate and verify the quality of the LLM-extracted rationales. Furthermore, an interesting and responsible direction forward would be the development of hybrid approaches that leverage LLMs for extracting rationales at scale and then employing human experts to verify the validity and quality of these rationales. This would also alleviate some of the concerns surrounding LLM hallucinations and biases in the LLM being propagated into the rationale extraction step.

\section{Limitations}




While our \textbf{SHIELD} framework shows promise in leveraging large language models to create interpretable hate speech detectors, several limitations need to be addressed. A inherent trade-off exists between the interpretability gained through LLM-extracted rationales and the accuracy of the resulting model, requiring further work to optimize this balance. In certain cases, the LLM may fail to identify coherent rationales, leading to incomplete or inaccurate explanations for the model's predictions. The choice of the LLM itself is also crucial, as powerful proprietary models like ChatGPT may not be accessible to all researchers, while open-source alternatives could potentially yield suboptimal performance. Our work currently uses ChatGPT for rationale extraction, but exploring the capabilities of different LLMs, including multilingual and domain-specific models, could provide valuable insights. Additionally, our framework may need adaptation to handle instances where the LLM cannot provide clear rationales, either through ensemble methods or by incorporating human feedback mechanisms to refine the extracted rationales.

\section{Ethical Considerations}
\subsection{Acknowledgment of the sensitivity and potential harm of hate speech}
We acknowledge that hate speech is a sensitive and potentially harmful topic that can perpetuate discrimination, marginalization, and violence against individuals or groups based on their race, ethnicity, religion, gender, sexual orientation, or other protected characteristics. We recognize the importance of addressing hate speech responsibly and with great care, as it can have severe psychological, emotional, and social consequences for those targeted. However, our work strives to better interpret and mitigate the use of hateful speech promptly by employing LLMs in an out-of-the-box manner leveraging their context-understanding capabilities in hate speech detection task.

\subsection{Commitment to responsible use and mitigation of potential misuse}
Our research focuses on leveraging the contextual understanding capabilities of large language models (LLMs) to automate the detection of hateful content, such as derogatory language, cuss words, and profanities, in the form of rationales across social media platforms. This aims to enable early-stage identification and mitigation of hate speech. We acknowledge the severity of the hateful examples used, which may potentially promote racial superiority, incite racial discrimination, or encourage violence against certain racial or ethnic groups – actions that are considered punishable offenses by law.  After a thorough evaluation, we have concluded that the benefits of using real-world practical examples to enhance the clarity and understanding of our research outweigh any potential risks or drawbacks associated with their inclusion.

\subsection{Ethical guidelines and principles followed}
In conducting our research, we adhere to established ethical guidelines and principles, such as those outlined by professional organizations and academic institutions. We have utilized publicly available datasets that are appropriately cited in our paper. We also strive to maintain transparency by clearly documenting our methods, data sources, and limitations.

\section*{Acknowledgements}

This work is supported by the DARPA SemaFor project (HR001120C0123), and the Office of Naval Research (N00014-21-1-4002). The views, opinions and/or findings expressed are those of the authors.

\bibliographystyle{acl_natbib}
\bibliography{references}




\end{document}